\documentclass{article}
\usepackage[table]{xcolor}
\usepackage{acra}
\usepackage{multirow}
\usepackage{color}
\usepackage{todonotes}
\usepackage{comment}
\usepackage{ragged2e}
\usepackage{siunitx}
\usepackage{graphicx}
\usepackage{multicol}
\usepackage{subfig}
\usepackage[english]{babel}
\usepackage[nottoc]{tocbibind}
\usepackage{url}

\title{A Sign Language Recognition System with Pepper, Lightweight-Transformer, and LLM}
\author{JongYoon Lim$^{1}$, Inkyu Sa, Bruce MacDonald$^{1}$, and Ho Seok Ahn$^{1}$\\ $^{1}$ \quad CARES, Department of Electrical, Computer and Software Engineering, University of Auckland \\ 
\{jy.lim, b.macdonald, hs.ahn\}@auckland.ac.nz, enddl22@gmail.com}

\begin{document}

\maketitle

\begin{abstract}
This research explores using lightweight deep neural network architectures to enable the humanoid robot Pepper to understand American Sign Language (ASL) and facilitate non-verbal human-robot interaction. First, we introduce a lightweight and efficient model for ASL understanding optimized for embedded systems, ensuring rapid sign recognition while conserving computational resources. Building upon this, we employ large language models (LLMs) for intelligent robot interactions. Through intricate prompt engineering, we tailor interactions to allow the Pepper Robot to generate natural Co-Speech Gesture responses, laying the foundation for more organic and intuitive humanoid-robot dialogues. Finally, we present an integrated software pipeline, embodying advancements in a socially aware AI interaction model. Leveraging the Pepper Robot's capabilities, we demonstrate the practicality and effectiveness of our approach in real-world scenarios. The results highlight a profound potential for enhancing human-robot interaction through non-verbal interactions, bridging communication gaps, and making technology more accessible and understandable.

\end{abstract}

\begin{figure*}
\center
\includegraphics[width=0.9\textwidth]{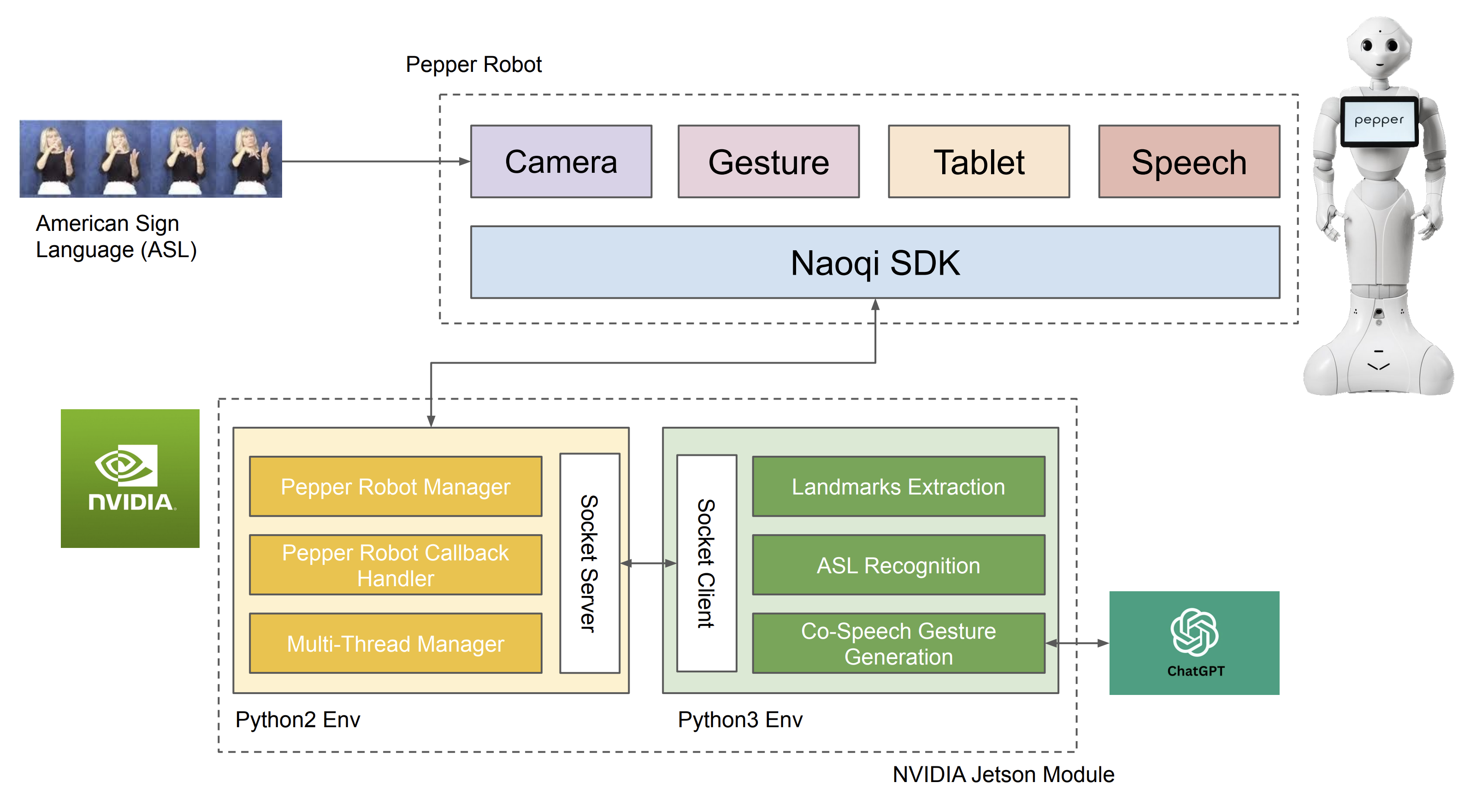}
\caption{System Overview: frames capturing signs from Pepper are conveyed to the Jetson module, where landmarks are extracted and relayed to the ASL Recognition model. Subsequently, Co-Speech Gesture outputs are derived from ChatGPT and transmitted back to Pepper, enabling the execution of corresponding gestures and dialogue.}
\label{systemoverview}
\end{figure*}

\section{Introduction}

Each day in the United States, approximately 33 infants are born with irreversible hearing loss \cite{CDC2010} , with around 90\% of these infants born to parents with average hearing ability and potentially lacking proficiency in American Sign Language (ASL) \cite{mitchell2004chasing}. The absence of sign language exposure places these infants in peril of Language Deprivation Syndrome, a condition defined by the absence of accessible, naturally acquired language within their critical language development period \cite{Hall2017-mj}. This syndrome has profound implications, affecting various life aspects, including relationships, education, and employment.

To ensure accessible learning of sign language and address the potential challenges of lack of language exposure, various platforms exist to facilitate the learning of sign language. \cite{suzgun2015hospisign} \cite{martins2015accessible}. Notably, multimodal platforms like robots have emerged as highly effective in language instruction, attributed to their interactive and adaptable learning settings \cite{uluer2015new} . These platforms can meet individual learning necessities and preferences, presenting a multifaceted approach to acquiring language that surpasses conventional instructional methodologies. Integrating Social Human Robots emerges as a pivotal solution \cite{zakipour2016rasa}. These robots are envisaged to mitigate the challenges inherent in learning sign language. By utilizing these advanced technologies, it is feasible to construct more inclusive and adjustable learning experiences, allowing a broader spectrum of individuals to communicate proficiently via sign language and consequently reducing the negative impacts of a lack of language acquisition.


However, recognizing sign language and generating human-like gestures in robotic systems is inherently computationally intensive and incredibly challenging for platforms with limited computational resources \cite{joksimoski2022technological}. The demand for real-time data processing, inherent to sign language recognition and natural gesture generation, necessitates high computational throughput and low latency \cite{sabyrov2019towards}. Additionally, deploying sophisticated machine learning algorithms, such as deep neural networks for feature extraction, recognition, and capturing temporal sequences, imposes an additional computational burden.

To address the previously highlighted challenges, we have created a comprehensive system for understanding sign language and making gestures, specifically designed for the Pepper robot. Our main contributions are outlined below:
\begin{itemize}
\item Sign Language Recognition: We developed a lightweight Deep Neural Networks (DNNs) model for understanding American Sign Language, optimized for systems with limited computing power.
\item Smart Interactions: We employed low-level motions and carefully designed prompts to enable Pepper to interact intelligently, producing appropriate and context-aware gestures using a Large Language Model (LLM) such as ChatGPT.
\item Complete Integration: We have built a fully integrated approach that combines these elements to enable social interactions between Pepper and humans, paving the way for more advanced human-robot interactions in the future.
\end{itemize}

\begin{figure*}
\center
\includegraphics[width=\textwidth]{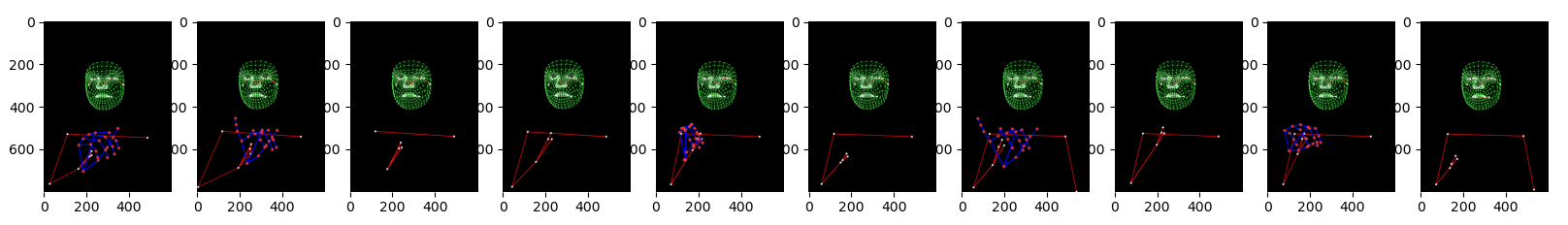}
\includegraphics[width=\textwidth]{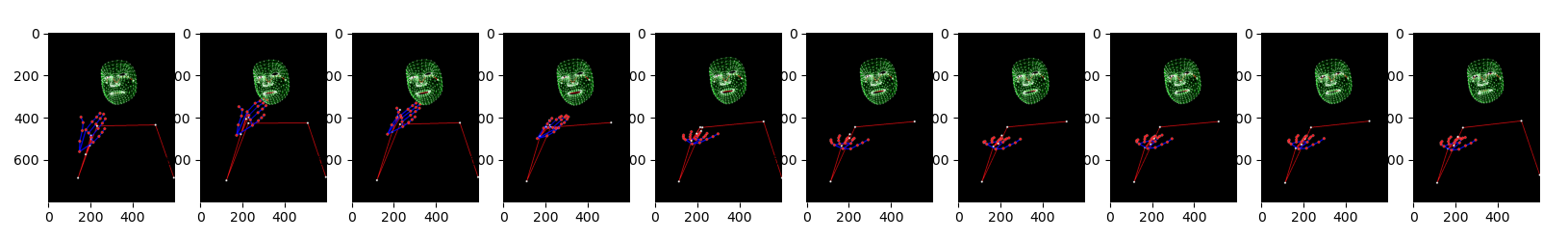}
\includegraphics[width=\textwidth]{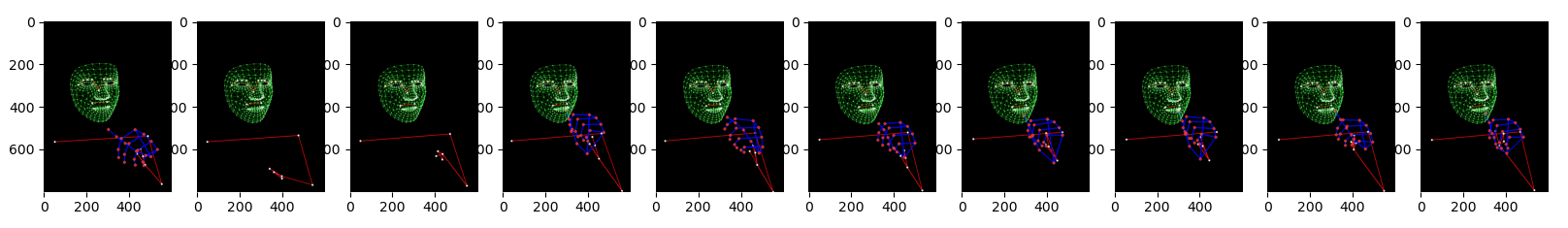}
\caption{Extraction of Landmarks Using Mediapipe: The top row represents the sign for the word 'same', the middle row depicts the sign for 'bad', and the bottom row illustrates the sign for 'nuts'.}
\end{figure*}

\section{Related Works}

\subsection{Sign Language Understanding using Deep Neural Networks}
Sign languages, natural languages conveyed through gestures and facial expressions, present unique challenges and opportunities in computer vision and AI. The evolution of DNNs has propelled advancements in the accurate recognition and translation of sign languages \cite{zuo2023natural} \cite{hu2021signbert} \cite{bohavcek2022sign}. Initial efforts in gesture recognition heavily relied on traditional computer vision techniques until the introduction of Convolutional Neural Networks (CNNs) \cite{rao2018deep}, which demonstrated enhanced proficiency in recognizing gestures by focusing on the spatial understanding of signs. Integrating Recurrent Neural Networks (RNNs) \cite{guo2018hierarchical} and transformer networks \cite{bohavcek2022sign} has proven effective in analyzing the sequential flow of sign gestures to capture the inherent temporal dynamics of sign language. Beyond gesture recognition, the capability of DNNs extends to end-to-end sign language translation, directly converting sign language to text or speech. Recognizing the multimodal nature of sign language \cite{kagirov2019method}, involving not just hand movements but also facial expressions and body posture, multimodal deep learning approaches have been advocated, amalgamating data from diverse sensors to refine recognition accuracy. The development of expansive datasets has been crucial in propelling this research, offering a diverse range of sign languages and signers for robust training and evaluation of DNNs \cite{li2020word} \cite{ronchetti2016lsa64} \cite{albanie2021bbc}. However, despite these advancements, challenges persist, including data scarcity, signer variability, and the complexities of recording non-manual signs. 

\subsection{Social Human-Robot Interaction (HRI)}
Research in human-robot interactions (HRI) has sparked interest in robotics and social sciences \cite{lemaignan2017artificial}, evolving from task-oriented interactions to socio-emotional exchanges resembling human-to-human interactions \cite{johanson2019effect}. Advancements in emotion recognition using deep learning enable robots to understand and respond to human emotions, facilitating seamless interactions. The development of robotic empathy has been crucial in fostering genuine human-robot connections, particularly in elderly care and education, where emotional support is vital \cite{gasteiger2022moving}. However, most current HRI focus on verbal communication, overlooking the significance of non-verbal cues like body language and facial expressions in enhancing interaction quality. The ability of robots to undertake perspective-taking improves collaborative work by considering human viewpoints and feelings. Applications of social HRI have yielded impressive results in fields like tutoring and counseling, underscoring the effectiveness of robots possessing socio-emotional skills. Nonetheless, achieving truly social and emotionally resonant HRI poses challenges, with areas such as the uncanny valley effect and the balance between robot autonomy and user control remaining key research domains.

\subsection{Large Language Model in Robotics }
The fusion of Large Language Models (LLMs) and robotics has sparked extensive research focusing mainly on prompt engineering, aiming to facilitate seamless human-robot interaction \cite{billing2023language}. Studies in prompt engineering within LLMs have paved the way for enhanced model responses, establishing foundational communication protocols between humans and robots. Researchers have demonstrated that integrating LLMs in robotic systems allows for the interpretation and execution of complex commands, emphasizing the critical role of optimal prompts \cite{Yu2023-fh}. Additionally, advancements in multimodal integration enable richer, context-aware interactions by combining visual and linguistic data. However, this integration has brought forth ethical concerns, such as bias and responsible deployment of technologies, necessitating meticulous consideration in their development and application. The practical applications of these integrations are extensive, with notable advancements in healthcare and education. Future research is directed towards refining prompt engineering techniques and developing more coherent interaction paradigms to effectively bridge the gap between natural language understanding and robotic responsiveness.

\subsection{Humanoid Robots in Education}
Humanoid robots, with their human-like appearance and dynamic interaction abilities, are increasingly being integrated into educational environments, from primary schools to universities, enhancing teaching methods and student engagement. These robots, as explored in studies \cite{leyzberg2014personalizing} and \cite{kennedy2016social}, serve as effective tutors, providing personalized, consistent, and adaptive learning experiences. They have proven particularly beneficial in language acquisition, offering immersive learning environments for students, especially in learning second languages. Additionally, their utility extends to special education, improving social interaction and focus for children with autism. In STEM education, humanoid robots act as educational tools for coding and robotics and as agents promoting problem-solving and critical thinking. Integrating humanoid robots necessitates understanding human-robot interaction, with studies \cite{scheutz201113} investigating the social dynamics, trust, rapport, and emotional bonding possibilities between students and robots. However, despite the multitude of benefits, challenges persist in areas like maintenance, teacher training, and balancing human and robot-led instruction, which are crucial to address for maximizing learning outcomes.

\subsection{Lightweight Deep Neural Networks in Robotcis}
Integrating lightweight DNNs with embedded systems like NVIDIA Jetson modules is an important advancement in robotics, drawing significant scholarly interest for its potential to enhance robotic capabilities. Research in this field has extensively focused on designing and optimizing lightweight DNNs to operate efficiently on resource-constrained systems \cite{ghimire2022survey}, with studies showcasing the deployment intricacies and advantages of utilizing NVIDIA Jetson modules for improved computational efficiency and power consumption in robotic applications. Significant work has been undertaken to integrate these optimized DNNs with robotic systems, enriching autonomous capabilities and enabling advanced real-time decision-making and environmental perception. Implementing these networks has allowed for real-time object detection and navigation, and advances in multi-sensor fusion have improved the robustness and accuracy of robotic perception modules. However, this domain faces challenges, especially in model optimization and resource allocation, with innovative solutions being proposed to overcome the limitations of embedded systems. Numerous application-specific developments have underscored the versatility and impact of lightweight DNNs in healthcare, agriculture, and industrial automation.

\section{Methodology}
The proposed system architecture(Figure \ref{systemoverview}) revolves around enabling Pepper Robot to interpret and interact using ASL. Users initiate communication through sign language, positioning themselves for clear visibility. The robot's camera sensor captures the user's gestures and postures, which are processed using Google's Mediapipe holistic tool to extract human body landmarks. These landmarks are relayed to a DNN model on an NVIDIA Jetson module, which identifies and classifies the signed word or phrase, considering the nuances of hand movements and facial expressions. The identified ASL is inputted into an LLM, like ChatGPT, which generates a corresponding verbal response and suggests appropriate gestures for Pepper Robot. These suggestions are converted into executable instructions using the Naoqi SDK, allowing Pepper Robot to respond to the user with verbal communication and corresponding gestures, offering an interactive and immersive experience.

The Isolated Sign Language Recognition Corpus (version 1.0) is an extensive compilation of approximately 100,000 videos featuring isolated signs. It encompasses hand and facial landmarks, created through Mediapipe version 0.9, and is articulated by 21 Deaf signers who predominantly use American Sign Language, employing a lexicon of 250 signs. The dataset contains columns denoting the frame number in the raw video, the type of landmark (which can be one of 'face', 'left hand', 'pose', 'right hand'), the landmark index number, and the normalized spatial coordinates of the landmark represented by [x/y/z].

\subsection{Sign Language Recognition}

\subsubsection{Dataset}
\begin{figure}[!htb]
\minipage{0.49\textwidth}
\includegraphics[width=\linewidth]{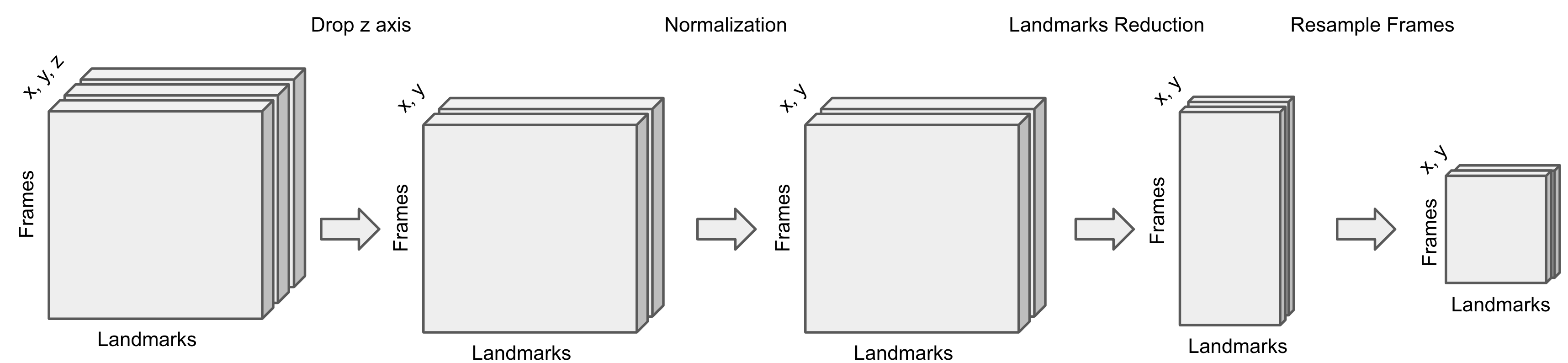}
\caption{Data Preprocessing. The input data frames undergo a series of transformations: dropping the z-axis, normalization, retaining only the required landmarks, and finally, resampling the frames.}
\label{datapreprocess}
\endminipage
\end{figure}

As shown in Figure \ref{datapreprocess}, only the coordinates of lips, hands, and arm pose are utilized in our approach. The landmarks are normalized using the mean and standard deviation of all landmarks, enhancing the model's overall performance. To further optimize performance, data augmentation plays a crucial role. Random resampling of the original length and random masking are employed for temporal augmentation. Additionally, spatial augmentation is implemented by applying horizontal flips and random affine transformations, which encompass scaling, shifting, rotating, and shearing.

\begin{figure}[!htb]
\minipage{0.49\textwidth}
\includegraphics[width=\linewidth]{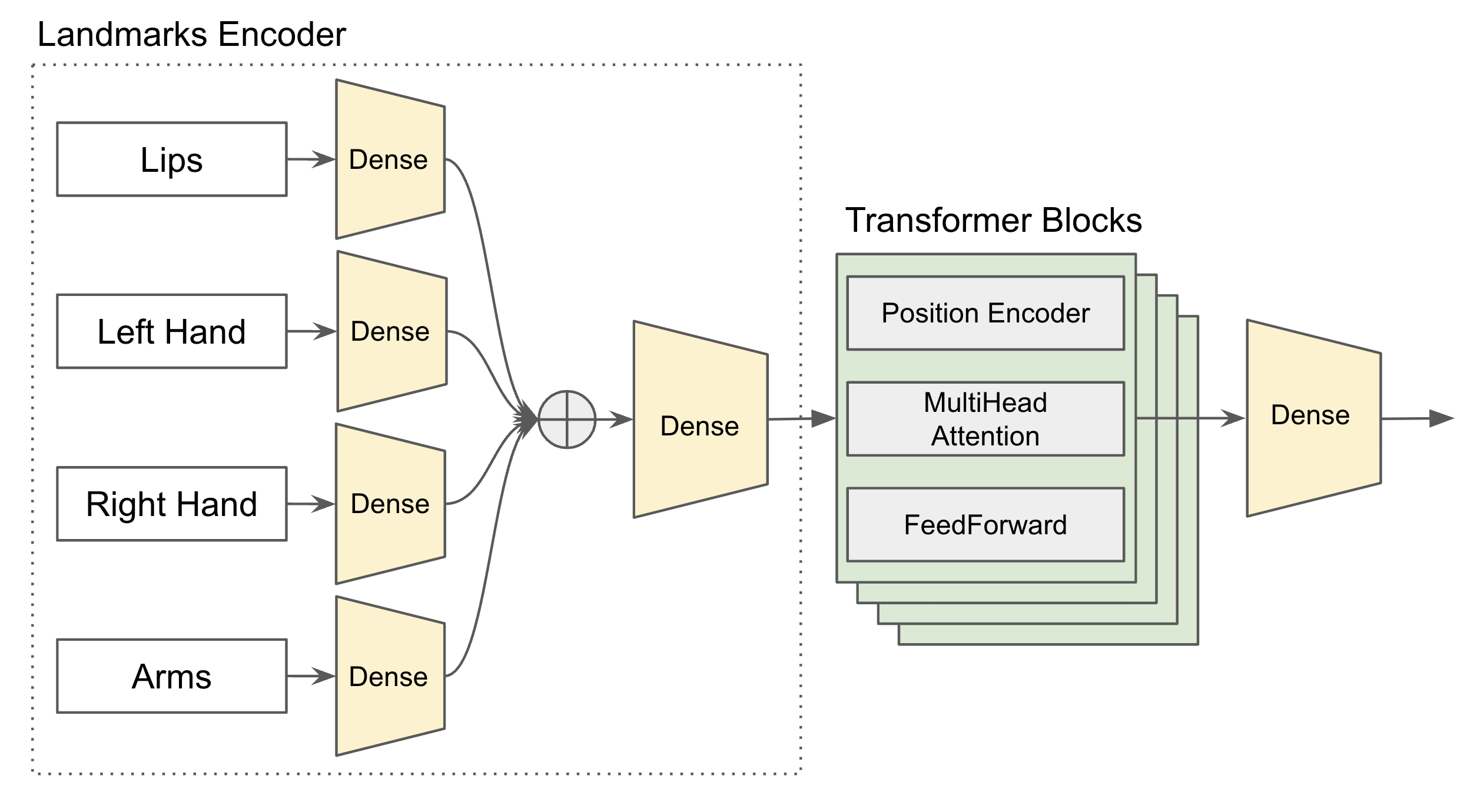}
\caption{Model Architecture. The preprocessed data is initially passed through a feature extractor and then combined. Subsequently, it is channelled through Transformer blocks before being fed into the classifier.}
\label{architecture}
\endminipage
\end{figure}

\subsubsection{Model}

This study employs a specialized model to extract features from landmarks. The initial phase of feature extraction involves the use of multiple dense layers, where each dense layer is succeeded by Layer Normalization and ReLU activation functions. The resulting extracted feature is then forwarded to four layers of a Transformer encoder, integral for processing sequential data and particularly potent for natural language processing tasks due to its self-attention mechanism.

\begin{figure}[!htb]
\minipage{0.49\textwidth}
\includegraphics[width=\linewidth]{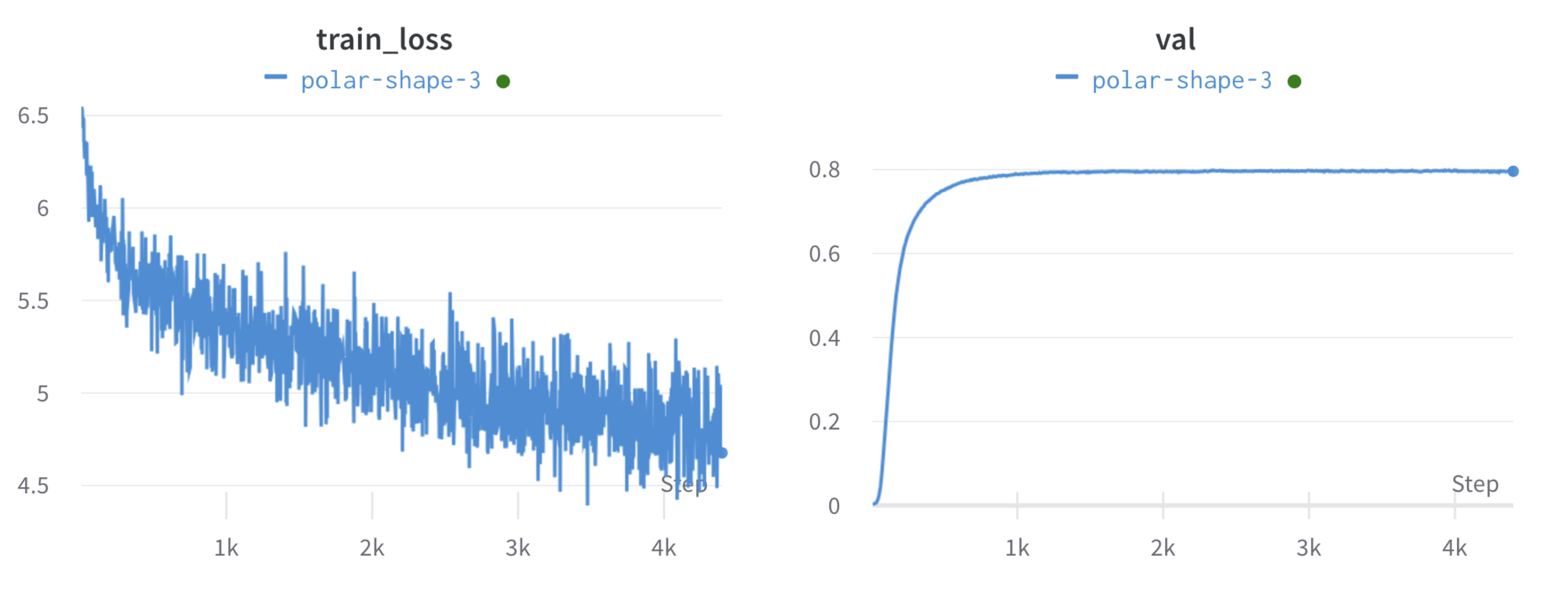}
\caption{Loss and Validation}
\label{lossval}
\endminipage
\end{figure}

The Transformer encoder processes the input sequence in this model architecture(Figure \ref{architecture}). It compresses the information into "context" or "memory," which a decoder usually would use to produce an output sequence in a typical Transformer model. However, the decoder is skipped in this research to ensure parameters and inference time efficiency. Instead, the output from the Transformer encoder layers is directly forwarded to a dense layer to obtain the logits for the classes, avoiding using an activation function in the final dense layer. This approach maintains model efficacy while optimizing computational resources and processing time. The model has a total of 2,562,970 parameters, which is relatively small, yet it still performs reasonably in recognizing ASL.

\begin{table}[h!]
\centering
\begin{tabular}{ |p{2cm}|p{3cm}|  }
 \hline
 &  Playtime (second) \\
 \hline
 mean & 4.09 \\
 std & 3.24 \\
 min & 0.51 \\
 25\% & 2.02 \\
 50\% & 2.90 \\
 75\% & 4.90 \\
 max & 24.4  \\
 
 \hline
\end{tabular}
\caption{ Play Time Statistics for the recorded 430 gestures executed by Pepper. }
\label{playtime}
\end{table}

\subsection{Co-speech gesture Dialogue generation using LLM}
Given the capabilities of Large Language Models (LLMs) in understanding and interpreting context within sentences, extracting emotions, sentiments, and other nuanced aspects of language, they serve as powerful tools for enriching interactions with humanoid robots like Pepper. LLMs can be instrumental in generating appropriate and meaningful gestures for Pepper, synchronized with its spoken subtitles, enhancing the overall communicative experience. 

\begin{table}[h!]
\centering
\begin{tabular}{ |p{2cm}|p{5.5cm}|  }
 \hline
 Gesture Tag   & Thinking \\
 \hline
 Description & The robot gently taps its head with its right hand, moving carefully and smoothly, like it’s deep in thought. \\
 \hline
 Playtime(s) & 2.17 \\
 \hline
 Moving Body Parts & Eyes, Neck, Right Arm, Right Hand \\
 \hline
\end{tabular}
\caption{A sample of Pepper's gesture descriptor. It includes a gesture tag, a human-authored description, its play time in seconds, and the specific robot body parts required to execute the gesture.}
\label{descriptor}
\end{table}

A two-step request(Table \ref{twostep}) to the model can be employed to leverage LLMs for integrating meaningful gestures. Initially, dialogue can be converted to speech using models like ChatGPT. Subsequently, the output from the first step can be prompted to incorporate gesture tags around specific words or sentences, creating a richer, more immersive interaction by aligning gestures with the spoken content. Providing a prompt to the LLM is crucial for generating natural and socially aware outputs. In the prompt instruction, we incorporate gesture descriptors(Table \ref{descriptor}) to convey more detailed information about Pepper's predefined gestures. The playtime statistics are displayed in Table \ref{playtime}.

\begin{table}[h!]
\centering
\begin{tabular}{ |p{8cm}|  }
 \hline
 \rowcolor{gray!30} First Step) Input to LLM\\
 \hline
 [INSTRUCTIONS] \newline
 \#\#\#\# \newline
 A signer accurately depicted a cloud with a 90\% accuracy rate. \\
 \hline
 \rowcolor{gray!30} First Step) Output from LLM \\
 \hline
 Great! You drew a cloud sign, but the weather today is really nice. Just look up at the sky. \\
 \hline

 \hline
 \hline
 \rowcolor{gray!50} Second Step) Input to LLM\\
 \hline
  [INSTRUCTIONS|Gesture Descriptors] \newline
  \#\#\#\# \newline
 Great! You drew a cloud sign, but the weather today is really nice. Just look up at the sky. \\
 \hline

 \rowcolor{gray!50} Second Step) Output from LLM\\
 \hline
 [Yes] Great! [/Yes] You drew a cloud sign, but [Excited] the weather today is really nice [/Excited]. Just [ShowSky] look up at the sky [/ShowSky].\\
 \hline
 
\end{tabular}
\caption{Table illustrating the two-step processing approach to generate Co-Speech Gesture using ChatGPT. The initial step utilizes the recognized word and its accuracy to generate a prompt with a specific INSTRUCTION. In the second stage, the returned output is processed using specific INSTRUCTION and Gesture Descriptors. The concluding output is text interspersed with gesture tags.}
\label{twostep}
\end{table}

\subsection{Deployment Model to NVIDIA Jetson module}
To deploy a trained model to the NVIDIA Jetson module, a transformation of the PyTorch model into TensorRT is essential. TensorRT, developed by NVIDIA, stands out as a high-performance deep learning inference library, fine-tuned to enhance the speed and efficiency of deep learning models during the inference phase. It is specifically designed to optimize and accelerate the deployment of models in environments like embedded systems, which is characteristic of the NVIDIA Jetson module. This conversion assures optimal utilization of the board’s resources and guarantees swift and efficient model inferences, making it highly suitable for embedded boards where enhanced performance and resource optimization are crucial.

\subsection{Communication between Pepper and Jetson module}
The Pepper robot operates using Python 2, while the Jetson module, assigned the task of interpreting sign language, utilizes Python 3. A socket network program establishes effective communication between Pepper and the Jetson module. This approach is founded on network protocols, typically TCP/IP, enabling data exchange between the applications running on Pepper and the Jetson module, which operate as different machines in the network. Socket programming is integral in this setup as it allows for creating scalable and robust network applications, providing a bidirectional communication link between the endpoints. This method is efficient, swift, and integrates seamlessly with other processes, ensuring smooth and responsive interaction between different system components.

\section{Results}
In the pursuit of bridging communication gaps using AI and robotics, this research has generated noteworthy findings. As we leveraged a confluence of technologies ranging from DNNs to Large Language Models, our empirical observations highlighted the strengths and challenges inherent in our approach. Our observations underscore a significant step forward in using sign language in human-robot interaction. While certain areas, such as depth prediction for landmarks extraction, require further refinement, the overarching results signify a promising foundation for future enhancements.

\subsection{ASL Recognition on Jetson module}
Our custom-developed ASL recognition model, optimized for the NVIDIA Jetson module, demonstrated a commendable accuracy rate. Upon testing, the model achieved an accuracy of 79.8\% as shown in Figure \ref{lossval}. This is particularly promising, considering the computational constraints of the Jetson module and the complexity inherent in recognizing the nuances of sign language.

\subsection{Mediapipe Holistic's Performance}
The Google Mediapipe holistic tool was employed for human body landmarks extraction. Our experiments indicated that the tool strongly predicted landmarks' x and y positions. However, its capabilities exhibited a limitation when it came to depth prediction. This aspect warrants further investigation and may necessitate supplementary techniques or sensors for robust three-dimensional understanding.

\subsection{ChatGPT's Multimodal Features}
One of the more intriguing observations came from deploying the ChatGPT LLM. ChatGPT showcased the ability to generate multimodal features. It was proficient in crafting dialogues while simultaneously generating a diverse array of gestures and emotions. This multi-faceted interaction potential reinforces the applicability of LLMs in human-robot interaction scenarios.

\subsection{Pipeline Integration}
Our integrated pipeline, which amalgamates multiple stages from ASL recognition to robot interaction, functioned seamlessly. The coherence and efficiency of the pipeline validate our architectural choices and implementations. Furthermore, the system is poised for scalability, indicating that it's ready to incorporate more meaningful experiments geared toward Social Human-Robot Interaction.

\section{Discussion}
In human-robot interaction, this research addresses the critical need for robots to comprehend and engage meaningfully with humans, especially those relying on ASL. A streamlined, resource-efficient model was developed for real-time ASL recognition, minimizing computational overhead in embedded systems. Incorporating LLMs allows for a deeper understanding of the intent, emotion, and context behind signs, enriching human-robot dialogues. The research's integrated pipeline epitomizes the collaboration of various AI technologies, establishing a foundation for socially aware AI interaction models enabling robots to relate to human users empathetically and intuitively. Future works aim at the system's expansion and refinement, especially in educational sectors, and improved ASL recognition, driving the vision of empathy and understanding robots in human interaction.

\section*{Acknowledgments}
This work was supported by the Science for Technological Innovation (UOAX2123, Developing a Reo Turi (Māori Deaf Language) Interpreter for Ngāti Turi, the Māori Deaf Community), and the Te Pūnaha Hihiko: Vision Mātauranga Capability Fund (Te Ara Auaha o Muriwhenua Ngāti Turi: The Journey of Muriwhenua Māori Deaf Innovation, UOAX2124) funded by the Ministry of Business, Innovation \& Employment (MBIE, New Zealand). Ho Seok Ahn* is the corresponding author.


\bibliographystyle{apalike}
\bibliography{references}

\end{document}